\documentclass[conference]{IEEEtran}
\usepackage[pdftex]{graphicx}
\usepackage{amsmath}
\usepackage[hidelinks]{hyperref}
\usepackage[utf8]{inputenc}
\usepackage{balance}
\usepackage{booktabs}
\usepackage{xcolor}

\ifCLASSINFOpdf
\else
\fi
\usepackage{stfloats}
\IEEEoverridecommandlockouts%
\IEEEpubid{\makebox[\columnwidth]{\color{red}Author pre-print. Paper accepted for publication at 14th IEEE eScience.} \hspace{\columnsep}\makebox[\columnwidth]{ }}

\begin{document}
\title{DeepDownscale: a deep learning strategy for high-resolution weather forecast}
\author{\IEEEauthorblockN{Eduardo R. Rodrigues, Igor Oliveira, Renato L. F. Cunha, Marco A. S. Netto}
\IEEEauthorblockA{IBM Research}
}

\maketitle

\begin{abstract}
Running high-resolution physical models is computationally expensive and essential for many disciplines. Agriculture, transportation, and energy are sectors that depend on high-resolution weather models, which typically consume many hours of large High Performance Computing (HPC) systems to deliver timely results. Many users cannot afford to run the desired resolution and are
forced to use low resolution output. One simple solution is to interpolate results for visualization. It is also possible to combine an ensemble of low resolution models to obtain a better prediction.
However, these approaches fail to capture the redundant information and patterns in the low-resolution input that could help improve the quality of prediction. In this paper, we propose and evaluate a strategy based on a deep neural network to learn a high-resolution representation from low-resolution predictions using weather forecast as a practical use case. We take a supervised learning approach, since obtaining labeled data can be done automatically. Our results show significant improvement when compared with standard practices and the strategy is still lightweight enough to run on modest computer systems.
\end{abstract}

\IEEEpeerreviewmaketitle%

\section{Introduction}

Numerical simulation has become the third pillar of science, in addition to
theory and experimentation. Models have become more complex over time and are
always in need of the most up-to-date computing infrastructure. Particularly,
resolution is one key factor that affects computing demand to execute such
models~\cite{iii2005computing}. For example, doubling the resolution of a
two-dimensional grid leads to an increase of four times in number of points to
be processed and an increase of the time resolution to meet the convergence
condition (CFL)~\cite{courant1967partial}.

In weather and climate simulations, one typically runs a large scale but coarse
resolution model and then, with the output of this model as input, runs a finer
resolution regional numerical model. That is called \textit{dynamical
downscaling}~\cite{wilby1997downscaling}. This procedure allows for finer
resolution in an area of interest without the computing cost and time of running
the large scale model with the desired resolution. However, the regional
numerical model is still a complex and computationally demanding software to
run.

Much research has been done in using Convolutional Neural Networks (CNN) to
reconstruct high-resolution data from low-resolution inputs. CNNs have been
applied extensively in imaging research. The key assumption is that it is
possible to recover a high-resolution image from a low-resolution input due to
high redundancy in the data. Here we investigate if the same assumption holds
for data generated from the execution of physical models. This approach has the
significant advantage of allowing one to use a supervised learning approach,
without the cost of manually labeling the data: one can run the physical model
with both resolutions (high and low-resolution) as much as their resources
allows to produce training data. Alternatively, one can use observational data
whenever available.

In this work, we propose a Convolutional Neural Network that takes as input
low-resolution weather data and interpolates it into a high-resolution output.
The features are the output of a set of global models for which we crop a region
of interest. The labels are observations in high-resolution. We compare with a
linear interpolation of the same set of input models, which is a standard
practice in meteorological science~\cite{wood2004hydrologic}. In addition, we
compare with a regression method and a high-resolution regional model. We used
weather forecast as a case study to apply our CNN-based strategy to
generate high-resolution data as it is a typical area that embraces
new models and strategies to address its computational
needs~\cite{lawrence2018crossing}.

The paper is divided as follows: in the next section, we review the
state-of-the-art, which is in both image processing and weather downscaling. In
Section~\ref{sec:contrib}, we present our contributions. In
Section~\ref{sec:arch}, we present our neural network architecture and reasons
for our choices. The following section describes the experiments we performed
and Section~\ref{sec:results} has the results of these experiments. Our final
remarks are found in Section~\ref{sec:conclusion}.

\section{Related work}
\label{sec:relatedwork}

The procedure of recovering a high-resolution image or video from a low-resolution counterpart is known as super-resolution (SR)~\cite{park2003super}.
Much research has been devoted to this topic in the recent
years~\cite{romano2017raisr,shi2016real,bar2018using}. Also, it has many
applications in diverse areas, such as satellite data~\cite{cheeseman1996super},
medical imaging~\cite{oktay2016multi}, and video
processing~\cite{altunbasak2000maximum}.

A typical assumption of super-resolution techniques is that $N$ images with different
perspectives are used to recover the high-resolution input. A common model is linear and takes
the form
\begin{equation}
    y_k = D B_k M_k x + n_k\text{,}
\end{equation}
\noindent
where $y_k$ is the $k$th image generated from the high-resolution input $x$, $M_k$ translates and
rotates the input, $B_k$ is a blur matrix, $D$ is a subsampling matrix, and $n_k$ is an additive
noise. In this model, not only is $x$ unknown, but $M_k$, $B_k$ and $D$ must be
estimated as well. Non-linear models have also been explored, as in the work by
He \emph{et al.}~\cite{he2007nonlinear}.

In order to estimate $x$ from $y_k$, there are several approaches, some based on
learning from data and some direct methods. In the direct methods, some examples
are: (1) nonuniform
interpolation~\cite{kim1990reconstruction,keshk2014performance}, frequency
domain methods~\cite{tsai1984}, and regularized approaches~\cite{lee2003regularized}.

Super-resolution techniques based on learning are very appealing, because one
can generate training examples easily. An (approximate) inversion procedure is
estimated by means of training tuples ($x$, $y_1$, \ldots, $y_N$). Translating,
rotating, blurring, and subsampling images provide examples ($y_k$) for the
untouched images (labels).

Deep learning models for super resolution are based on ideas similar to
auto-encoders~\cite{hinton2006reducing}. Auto-encoders are feedforward neural
networks whose purpose is to reconstruct their inputs. They have two parts, the
encoder and decoder. Usually, auto-encoders have a bottleneck between the
encoder and the decoder, and the weights of the bottleneck are the
representation the user seeks. This representation is intended to be an efficient
code for the input. In the super-resolution setting, the encoding part is
equivalent to the process that generates the low-resolution image, the code
represents the low-resolution image, and the decoder is the procedure the user
is seeking to train.

Shi \textit{et al.}~\cite{shi2016real} showed a CNN capable of real-time 1080p
super-resolution on a single GPU\@. They developed an architecture that extracts
features in the low-resolution space. Moreover, they introduced
a sub-pixel convolution layer that learns an array of upscaling filters to
upscale the final low-resolution feature maps into the high-resolution image.
They assumed, however, a known noise function, that may not be the case in the
physical setting.

In weather and climate, the efforts for dynamical downscaling data normally make
use of higher resolution regional models nested within coarser global models.
Boundary conditions are provided by the global models while the physics in the regional
models are tasked with the simulation of the sub-grid processes that are not
represented in the coarser model. Yang \textit{et al.} \cite{yang2016biases} performed
simulations with the LMDZ4 model at a 0.6$^\circ$ $\times$ 0.6$^\circ$ resolution nested
in 3 models with resolutions varying from 1.125$^\circ$ $\times$ 1.125$^\circ$ to 2.8125$^\circ$
$\times$ 2.8125$^\circ$, successfully reproducing precipitation and temperature
patterns over the domains analyzed. Even the model with the lowest resolution showed
significant improvement after dynamical downscaling.

Heikkilä \textit{et al.} \cite{heikkila2011dynamical} performed dynamical downscaling of the ERA-40
reanalysis dataset using the regional model WRF and two nested domains: a 30km-resolution
domain over North Atlantic and a 10km-resolution domain over Norway. Spectral nudging
was applied to guarantee the regional model circulation would be consistent with the
forcing reanalysis data. The finer resolution domains were able to reproduce the extreme
precipitation events in the North Atlantic and to reduce regional temperature biases
over Norway.

Weather forecast \cite{hov2017five,bauer2015quiet,mizielinski2014high} is not
the only area that demands optimization to reach high resolution results.
Biology~\cite{groen2013flexible,hetherington2007addressing} and
medicine~\cite{neubert2012automated} are examples of areas that require
simulations of physical models in high resolution and could benefit from the
findings provided in this paper.

\section{Contributions}
\label{sec:contrib}

The contribution of this paper is twofold, (1) a learning-based strategy to fuse physical simulations to produce a high-resolution output applied to weather forecast, and (2) a comparative analysis of the limits of super-resolution approaches to downscale.

In our current approach, we use a feed-forward neural network to associate input patterns to high-resolution
output. As we will show with the results of our experiments, this strategy can indeed capture the redundancies
in the input and produce a forecast that is significantly better than standard procedures. Consequently,
these results show that even without memory, the network can still find relations that cannot be attributed
to known noise as in the image processing context.

In addition, our proposed initialization (discussed in Section \ref{sec:arch}) allows the use of deeper
neural networks that better approximate the regression function we are seeking. This works similar to the
skip connections of the Resnet architecture~\cite{targ2016resnet}.

Moreover, we show one usage for locally connected layers that allows for local specific pattern matching.
This is done in a context that a fully connected would not be possible, because of the memory requirements
of these layers and the output size of our network.

Finally, even though we have tested the strategy only with a weather application, we expect that it generalizes to other
applications. For example, Finite Difference Methods that are run routinely over the same domain can have the
regularities that the Neural Network architecture would capture. These regularities would allow one to run low-resolution
in the region of interest and obtain a higher resolution, even if a higher resolution simulation would be run posteriorly.
\section{Architecture}
\label{sec:arch}

\begin{figure*}[!th]
    \centering
    \includegraphics[width=.8\linewidth]{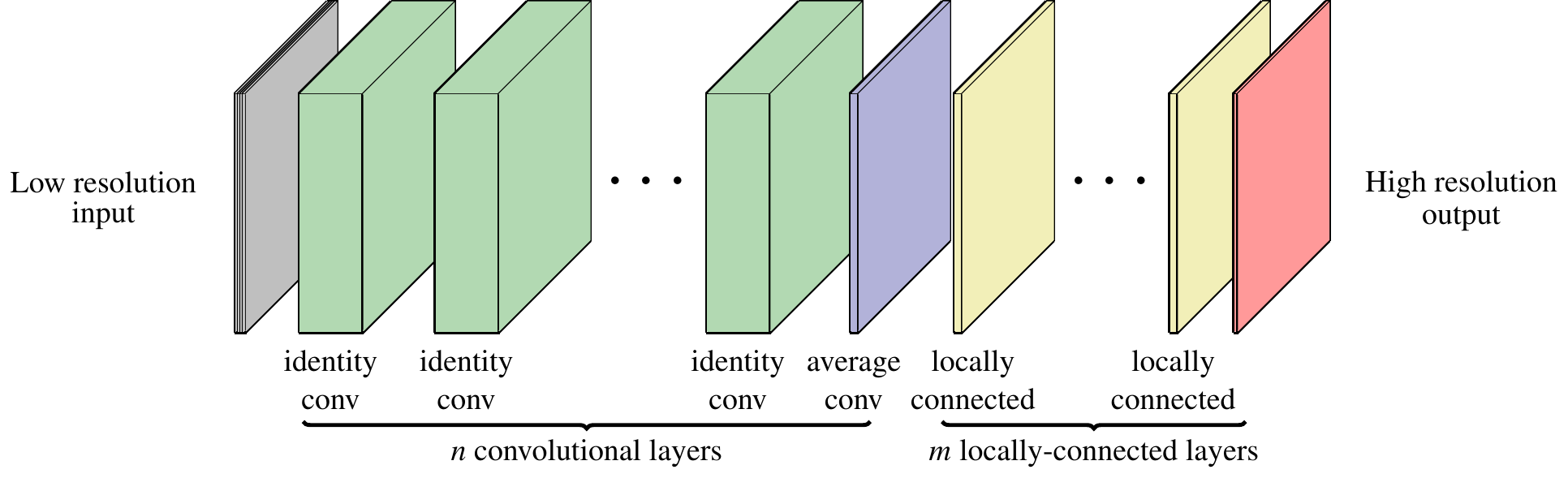}
    \caption{%
      Neural Network Architecture. The architecture is configurable and supports
      $n-1$ identity convolutions, each with eight output filters, represented in
      green in the Figure. The last convolution is initialized as an average and
      generates a single-channel output. After the convolutions, $m$
      locally-connected layers follow, shown in yellow. The low resolution
      inputs are shown in gray and the high-resolution output is shown in red.
    }\label{fig:arch}
\end{figure*}

In this section, we discuss our neural network architecture and the reasons for the choices we made.
In order to evaluate the impact of each of these decisions, we ran controlled experiments and present
results in the following sections.

The basic architecture we developed is presented in Figure \ref{fig:arch}. The input is a set of
$n$ low-resolution forecasts from $n$ different weather models. Each input is interpolated so that they
have the same horizontal dimensions. In our experiments, the horizontal dimension is the upscaled version
of the model. One could use a periodic reshuffling~\cite{shi2016real} and keep all the inputs in the
low-resolution dimensions and save memory. However, in our experiments we used the upscaled dimensions.
All input from the different weather models are then stacked together into a single volume of depth $n$.

The input volume is fed into a series of convolutions which are initialized with an approximate delta function up to
the last but one layer:

$$ f^l(x;W_l,b_l)=\phi(W_l*x+b_l)=\phi(\hat{\delta}*x) \simeq x $$

\noindent
the approximate delta function $\hat{\delta}$ is equal to one at $t=0$, but rather than being zero elsewhere, it is initialized
to very small random numbers in order to break symmetry. This needs to be done, because the pure delta function
has still many symmetries that would prevent backpropagation to compute distinct updates. The activation function ($\phi$) is
the rectified linear function (Relu).

This approximate delta layer can only produce the same activation depth as the input. In order to allow deeper
activations, we used a stack of $\hat{\delta}$:

$$ f^l(x;W_l,b_l)=\phi(W_l*x+b_l)=\phi([\hat{\delta}_1,... ,\hat{\delta}_m]*x) \simeq [x_1,... ,x_m] $$

\noindent
where $m$ is the number of copies of the input. This means that the activation depths can only be multiples
of $n$. This restriction can be relaxed by making partial copies of the inputs or adding extra intermediate
outputs that are zeroed by initial parameters down the network. However, in our experiments we kept the
number of activations just as multiples of $n$.

In the last layer, we initialize the weights with uniform distribution in the depth direction (so this layer
takes an average along the depth). Effectively, this initialization takes an average of the input models
as the final result. For this particular model, this is similar to what the Resnet architecture
\cite{targ2016resnet} would do with small weights, i.e. it will let the input to cross the network unchanged.
Similarly to the Resnet architecture, the optimization procedure adjusts the weights to improve
the response if it is possible to do so. This strategy allows for very deep neural networks without
suffering from premature overfitting.

\begin{figure}[h!]
    \centering
    \includegraphics[width=.35\textwidth]{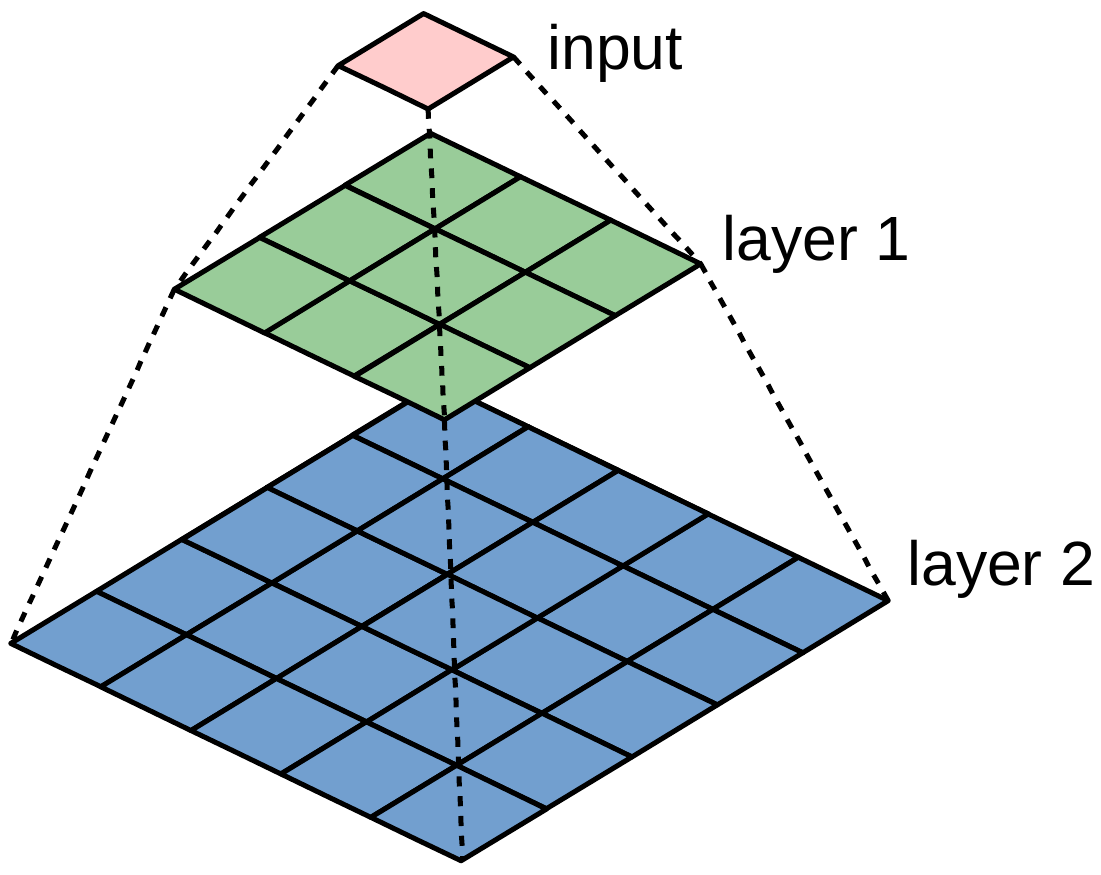}
    \caption{%
      View of the input after one and two 3x3 convolutions
    }\label{fig:ext}
\end{figure}

We used a cross-validation approach in order to select the number of convolutional layers. The best network
was very deep, with more than 40 layers. This fact appears to indicate that, as in other visual recognition
tasks, the number of layers substantially benefit from very deep networks. One possible explanation is
that the later convolutions need to have a larger view of the domain. That is, the constant receptive
field can only reach to a limited extension, the additional layers
allows that extension to expand as illustrated in Figure \ref{fig:ext}.

The extension viewed by a single activation can be increased by larger filters. One could, therefore,
increase the filter size and have the same extension and a shorter architecture. However, only extending
the view may not keep the same performance, since this strategy reduces the number of non-linearities and
consequently the expressiveness of the network. We investigated the relation between
view extension and network depth and concluded that simpler and more numerous layers were better for our
application.

The last locally connected layers are intended to capture patterns that are locality-specific. This is not
typically needed for neural networks in general and, in fact, is usually avoided. That is because for classification
and regression one seeks filters that activate when they find visual features irrespective to their location.
For example, a picture can be classified as a certain object even when that object has never appeared in the
given position in the training set. In our case, however, location is fixed and represents specific positions
in the domain. Consequently, we want to learn the particular influence of location in the forecast. For
example, topography plays a role in precipitation and can help refine the output. One could use a fully connected
layer to achieve the same result. However, this strategy would require a very large memory footprint. For instance,
the weighs of the fully connected layer would require $O(N^4)$, where $N$ is the horizontal dimension of the domain.

Lastly, the cost function used is simply the quadratic loss:
$$ L = \frac{1}{I*J*K} \sum_{i,j,k}^{I,J,K} ||y_{i,j,k}-\hat{y}_{i,j,k}||_2^2\text{,} $$
where $I$ is the number of examples, $J$ and $K$ are the two horizontal dimensions, and $y$ and $\hat{y}$ are
label and prediction respectively. In addition to the quadratic loss, we applied L2 regularization to the weights:
$$ L \mathrel{+}= \lambda \sum_l{W_l^2} \text{,}$$
where $\lambda$ is sought through cross-validation and $W_l$ is each one of the layer weights. This neural network
performs multiple regression, one for each forecast point.
\section{Experiments}
\label{sec:exp}

In order to evaluate the benefits of the proposed neural network, we made predictions and compared them to
standard procedures of downscale. Specifically, we used precipitation data. We compare our results
with an ensemble mean, a high-resolution local weather model and a linear regression ensemble model. In this section, we
describe in details the training procedure, the methods with which we compare the neural network
and, in the next section, we show the results.

The metric used to evaluate each downscale approach was the root of the cost function, i.e. the Root Mean Squared Error (RMSE).
This is appropriate since it measures the error in the units used in the forecast, and it allows us to evaluate how much
error on average each method has for each grid point.

We used as input data from the Coupled Model Intercomparison Project Phase 5 (CMIP5), a framework for comparison
of general circulation models and one of the main basis for the Intergovernmental Panel on Climate change
(IPCC) Assesment Reports \cite{taylor2012overview}. Specifically, we used the historical simulations,
from January 1st 1981 to December 31st 2005 of the models \textit{ccsm4}, \textit{cgcm3}, \textit{cm3} and \textit{cm5a}. Our decision
to use CMPI5 was its wide availability and its long continuous runs with the same model parameters.
The particular models were chosen because of the range of resolutions (Table \ref{tab:inputres}).

The labels could have been a high-resolution model output. In that case, one could generate input
data and labels from the same model. Specifically, low-resolution data would be used as input and high-resolution output as label. That procedure would permit an automatic supervised learning strategy, that
does not require human labeling, but incurs the cost of running (possibly many times) both low and
high-resolution models. Here, instead, we use observations as labels from the Chirps dataset \cite{funk2015climate},
whose resolution is 0.25 degrees.  Chirps has more than 30 years of precipitation data spanning $50^\circ$S-$50^\circ$N all longitudes.
This dataset incorporates both satellite and weather station data.

\begin{table}[!h]
\begin{center}
    \caption{%
        Resolution of input models.
    }\label{tab:inputres}
    \centering
    \def\arraystretch{1.05}
    \begin{tabular}{lc}
        \toprule
        model   & resolution (degrees) \\
        \midrule
        \textsc{ccsm3} & 1.25 \\
        \textsc{cgcm3} & 1.25 \\
        \textsc{cm3}   & 2.50 \\
        \textsc{cm5a}  & 3.75 \\
        \bottomrule
    \end{tabular}
\end{center}
\end{table}

In order to train the model, we first divided the available data into three sets: (1) training,
with forecasts and observations from January 1st 1981 to December 31st 2003, (2) validation,
January 1st 2004 to December 31st 2004, and (3) test, January 1st 2005 to December 31st 2005.
The validation set was used to choose the hyperparameters and the test set was used only to
report the final performance of the model. Since the data is temporal, it is of utmost importance not
to simply randomize and split this data, since this may lead to a biased test performance.

The first step in the training procedure was to tune the following set of hyperparameters: learning rate,
regularization coefficients, number of convolutions, and number of locally connected layers. The
learning rate was chosen randomly from a log-uniform distribution between $10^{-4}$ and $10^{-1}$.
The regularization coefficient was chosen from a wider distribution (log-uniform between $10^{-5}$
and $10^{1}$), to encourage the network explore more regularization values. That is because we
believe that there are too many parameters for the size of our training data.

We adopted the standard procedure of computing the mean and standard deviation of the training set,
and, with those values, centering and normalizing training, validation and test set. We do not use
dropout. The mini-batch size was set to 16. The total input has 8389 points, each of which with 4
low-resolution forecasts. The validation and the test set have 359 points each.

We used a single GPU GTX1080TI, which has 11GB of memory. Our intention was to have a system much
cheaper than those required to run the downscale typically done in weather forecast centers. Moreover,
most of the time was spent training the network, that is done only once (one may incrementally update
the model however). The evaluation is even cheaper to execute.

The results of our model were compared with the ensemble mean of the input models. This is a simple
method, but is typically better than control forecasts by most standard verification metrics \cite{no20121091}.
That is because it smoothes out outlier predictions. This can be particularly good if the
errors of the individual models are uncorrelated, in which case the errors are canceled. The method is
an instance of what has been called ``wisdom of crowds'', in which crowds (or ensemble of predictors) can
make better decisions compared to those of isolated individuals.

The ensemble mean has the advantage that it does not require any training. However, a consistently bad
model can reduce the accuracy of the ensemble. A simple approach to reduce this issue is to use linear regression. We
used both the training and validation sets as defined previously to train the model:

$$
\hat{Y}=AX+b
$$

\noindent
where $A$ and $b$ are the weights and biases of the linear model. We train this
model with the regular least squares method, since the amount of data is not
large.  This model will perform a linear combination of the ensemble $X$.

Finally, we compare our model with a high-resolution weather model from the
CORDEX project \cite{jones2011coordinated}. CORDEX is a multi-institution effort
to produce high-resolution climate simulations via dynamical downscaling. CORDEX
proposes several regional domains across the globe, including a 0.44$^\circ$
resolution domain over South America that matches the domain used in this paper.
The simulation used here is a downscaling experiment of the ``historical'' runs
from CMIP5 \cite{taylor2012overview} that runs from 1950-2005. Here, the daily
precipitation for the 2001-2005 period was used to validade the downscaling
method. CORDEX represents the typical computationally expensive method to produce
high-resolution climate data, since its simulations may take up to 10.6 hours to
run for domains of similar dimensions and resolution \cite{cordexSMHI}.

\section{Results}
\label{sec:results}

In this section we present a comparison between the proposed neural network and each one of the three
traditional alternatives for downscale, i.e. ensemble mean, multiple linear regression model, and a
regional model. This comparison was done with a test set not seen by any model and subsequent to both
the validation and training sets, that were used to train the neural network, but also to train the
multiple linear regression model.

We tested seven days forecast for rain. This configuration is appropriate for a number of applications.
For example, farmers could use the forecast to decide seeding and pesticide application. The first comparison
is shown in Table \ref{tab:error}. This table shows the root mean square error (RMSE) of the neural network and
the three alternatives compared with the observed rain.

\begin{table}[!h]
\begin{center}
  \caption{root mean square error in the test set.}\label{tab:error}
  \def\arraystretch{1.05}
  \begin{tabular}{ l r }
    \toprule
    Model & \textsc{rmse} (mm) \\
    \midrule
    Neural network & 24 \\
    Ensemble mean & 38 \\
    Linear regression & 27 \\
    Regional model & 33 \\
    \bottomrule
  \end{tabular}
\end{center}
\end{table}

On average, the neural network achieved the best performance. It is interesting to notice that
the regional model was {\em not} the second best. There is a number of reasons to explain this
fact. One possibility is the inherently biases these types of models may have. Since we used
raw forecasts, this may be one reason for the poor performance. In addition, the neural network
was trained with the observation itself, which has helped it to perform better.

\begin{figure}[t!]
    \centering
    \includegraphics[width=1.\linewidth]{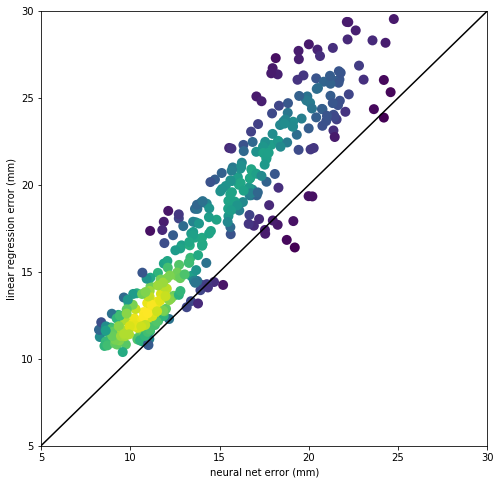}
    \caption{%
      Comparison between linear regression and Neural network error
    }\label{fig:perfcompare}
\end{figure}

In addition, the multiple linear regressor had a large intercept coefficient, as it can be seen
in Table \ref{tab:lincoef}. This suggests that the models under estimate the precipitation systematically. Consequently,
a bias adjustment will account for most of the improvement compared to the ensemble mean.
However, compared to the neural network, the multiple linear regression still systematically
has a worse error, as it can be seen in Figure \ref{fig:perfcompare}. In this figure, each
forecast error of both the neural network and the linear regression is paired. The errors
in the linear regression are almost always larger than the neural network errors.

\begin{table}[h!]
\begin{center}
  \caption{Multiple linear regression coefficients.}\label{tab:lincoef}
  \def\arraystretch{1.05}
  \begin{tabular}{ l r r }
    \toprule
    & Coefficient & Standard Error\\
    \midrule
    \textsc{ccsm3} & 0.21 & 2e-3 \\
    \textsc{cgcm3} & 0.10 & 7e-5\\
    \textsc{cm3} & 0.30 & 8e-5\\
    \textsc{cm5a} & 0.12 & 8e-5\\
    intercept & 7.90 & 5e-5\\
    \bottomrule
  \end{tabular}
\end{center}
\end{table}

\begin{figure}[b!]
    \centering
    \includegraphics[width=1.\linewidth]{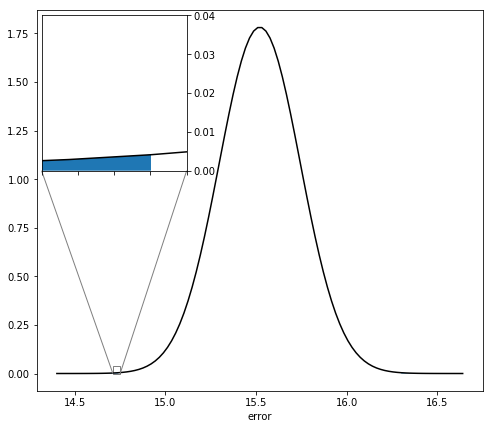}
    \caption{%
      Confidence analysis
    }\label{fig:analysis}
\end{figure}

In order to evaluate the significance of the improvement observed with the neural network, we
computed the probability of making daily prediction with the test set assuming that the network
was better than the interpolation just by chance ($H_0$). That is, the null hypothesis is that the
neural network is no better than the linear interpolation. The computed probability was less than 0.1\%.
This is illustrated in Figure \ref{fig:analysis}.

\begin{figure}[t!]
    \centering
    \includegraphics[width=1.\linewidth]{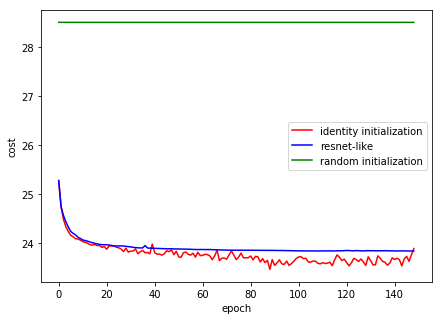}
    \caption{%
      training cost comparison
    }\label{fig:resnet}
\end{figure}

\begin{figure*}[t!]
    \centering
    \includegraphics[width=1\linewidth]{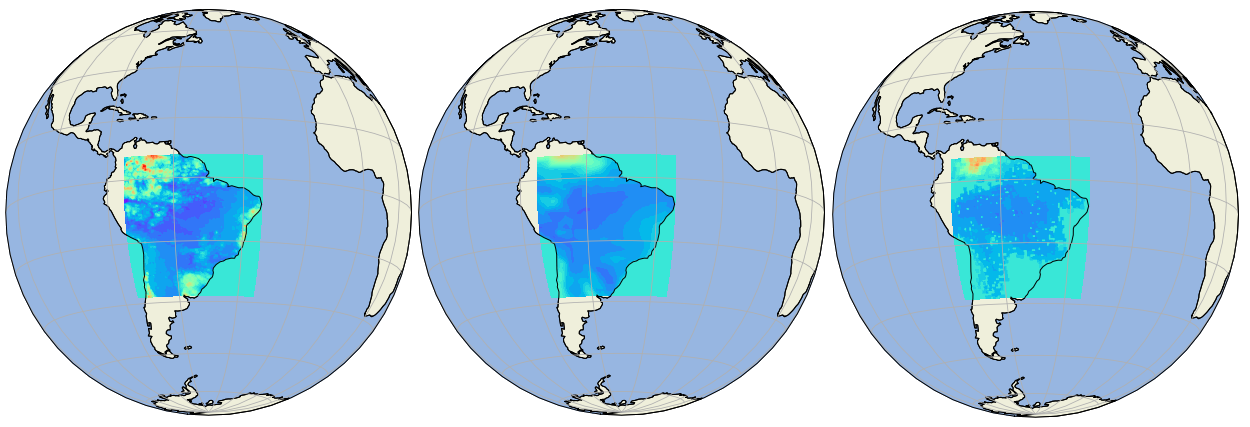}
    \caption{%
     Comparison between observed weather (left), best interpolated forecast (center), and network network output (right).
    }\label{fig:pred0}
\end{figure*}

However, what we really want is the probability of the neural network being better than the interpolation
given the experiments, i.e. $P(H_1|O)$, and what we have is $P(O|H_0)$. There is also the assumption that
the only hypotheses are $H_0$ and $H_1$, but this is reasonable and means that we are comparing whether or not
the best conventional strategy is better than the neural network. Still, the $99.9\%$ change of being better
is only true if the probability of the null hypothesis is the same as the probability of 
the observation (results of the experiments):

\begin{align*}
P(H_1|O) &= 1 - P(H_0|O) \\
&= 1 - \frac{P(O|H_0) P(H_0)}{P(O)}
\end{align*}

In order to compute $P(H_1|O)$ accurately, we would need to have (or estimate) the marginals $P(H_0)$ and $P(O)$. This
is not an easy task, since we have no access to the distribution of observations and we have limited observation
samples. In this setup, however, we can evaluate the limits to which $P(H_0)/P(O)$ can go so that we accept $P(H_1|O)$ as
likely.

Consequently, from the results we obtained, the ratio between $P(H_0)$ and $P(O)$ can be as high as 50 and still the significance will be
larger than 95\%, i.e. the experiment results can be 50 times less likely than the null hypothesis and
still the significance will be larger than 95\%.

We also compared our neural network with a Resnet-like architecture. In this model, the same structure as
described previously (see Figure \ref{fig:arch}) is enhanced with ``shortcut connections'', i.e. every two
layers, the input is copied verbatim to the output. The weights are initialized with small random values,
following the standard practice.

We obtained similar results to our architecture with the Resnet-like model. This suggests that the identity
initialization places the weights close to the solution that the resnet will find. This is not what typically
happens in other applications. For example, in image classification the gradient decent method
could in principle find the identity function in deep neural networks. However, this is not what happens
as reported by He \textit{et al.} \cite{he2016deep}. In Figure \ref{fig:resnet}, we compare the cost during
training for our architecture, resnet and our architecture with random initialization. As it can be seen,
out architecture has a similar, although noisier, cost evolution.

Finally, we compared the prediction itself of both the best interpolated model and the neural network output.
Figure \ref{fig:pred0} shows the observation, the best interpolated forecast, and the output of the neural network output. As it can be seen, the neural network result in this particular forecast has more details in the storm over the Amazon.

\section{Conclusion}
\label{sec:conclusion}

Downscaling is an important procedure for weather and climate applications in which coarse resolution forecasts are refined to meet a desired resolution. Many users rely on
downscaling results to make decisions in many disciplines. A typical form of this procedure is
known as dynamical downscale in which a high-resolution regional model is run with low-resolution data from another model as input.
However, running regional models in the required resolution is very costly. 

On the other hand, much research has been done to improve resolution of images (and video) in
computer science, in what is known as super-resolution. Typically, these techniques rely on the
fact that much information is redundant and a high-resolution image can be recovered from the
low-resolution input. However, most of the literature assume a known noise/error function. This is
not the case of downscaling, in which the reverse mapping between the high-resolution to the low-resolution forecast is completely unknown. Still, in
this paper we investigate if the super-resolution technique could be used to perform downscale of weather models and compared with standard procedures.

Our proposed strategy is based on Deep Neural Networks. This approach has a major advantage that the super-resolution procedure
is learned from data in a supervised learning fashion. Moreover, there is no need to manually labeling the data, since one can always run the model
in both resolutions to generate training examples - even though there is a computational cost associated with this procedure - but also it is 
possible to use observations as labels. From our experiments, we observed significant improvement of the proposed strategy compared
with standard downscale procedures. Moreover, the strategy is cheap enough to run in a single GPU system, and even training can be run
on that system.

Finally, we expect our strategy to apply to other applications where multiple models are run over the same domain. A particular example is soil
moisture, that can benefit high-precision agriculture. In this application, multiple soil moisture models are available to a given region. However,
some high-resolution data are only available with a few days delay. One could use our strategy to approximate the high-resolution data from the
low-resolution but updated data. Still, one would need to deal with the time delay that may require a model with memory.
 
\bibliographystyle{IEEEtran}
\balance
\bibliography{ref}
\end{document}